\documentclass[a4paper]{IEEEtran} 
\IEEEoverridecommandlockouts
\usepackage{amsmath,amssymb,amsfonts}
\usepackage{algorithmic}
\usepackage{graphicx}
\usepackage{xcolor}
\usepackage{textcomp}
\usepackage{xcolor}
\usepackage[
        backend=biber,
        style=ieee,
        sortcites=true,
    ]{biblatex}
\begin{document}

\title{Enhancing Microgrid Performance Prediction with Attention-based Deep Learning Models}

\author{Vinod Kumar Maddineni$^1$, Naga Babu Koganti$^2$ and Praveen Damacharla$^3*$\\
$^1$Power Systems Engineer, Schneider Electric, CA, USA\\
$^2$Power Electronics Engineer, KineticAI Inc., MI, USA\\
$^3$Research Scientist, KineticAI Inc., The Woodlands, TX, USA\\
$^*$praveen@kineticai.com
}

    
        
  
\maketitle

\begin{abstract}
In this research, an effort is made to address microgrid systems' operational challenges, characterized by power oscillations that eventually contribute to grid instability. An integrated strategy is proposed, leveraging the strengths of convolutional and Gated Recurrent Unit (GRU) layers. This approach is aimed at effectively extracting temporal data from energy datasets to improve the precision of microgrid behavior forecasts. Additionally, an attention layer is employed to underscore significant features within the time-series data, optimizing the forecasting process. The framework is anchored by a Multi-Layer Perceptron (MLP) model, which is tasked with comprehensive load forecasting and the identification of abnormal grid behaviors. Our methodology underwent rigorous evaluation using the Micro-grid Tariff Assessment Tool dataset, with Root Mean Square Error (RMSE), Mean Absolute Error (MAE), and the coefficient of determination (r2-score) serving as the primary metrics. The approach demonstrated exemplary performance, evidenced by a MAE of 0.39, RMSE of 0.28, and an r2-score of 98.89\% in load forecasting, along with near-perfect zero state prediction accuracy (approximately 99.9\%). Significantly outperforming conventional machine learning models such as support vector regression and random forest regression, our model's streamlined architecture is particularly suitable for real-time applications, thereby facilitating more effective and reliable microgrid management.
\end{abstract}

\begin{IEEEkeywords}
Load forecasting, Microgrid management, CNN, GRU, Attention.
\end{IEEEkeywords}

\section{Introduction}
\begin{figure}[!t]
\centering
\includegraphics[width=0.7\columnwidth]{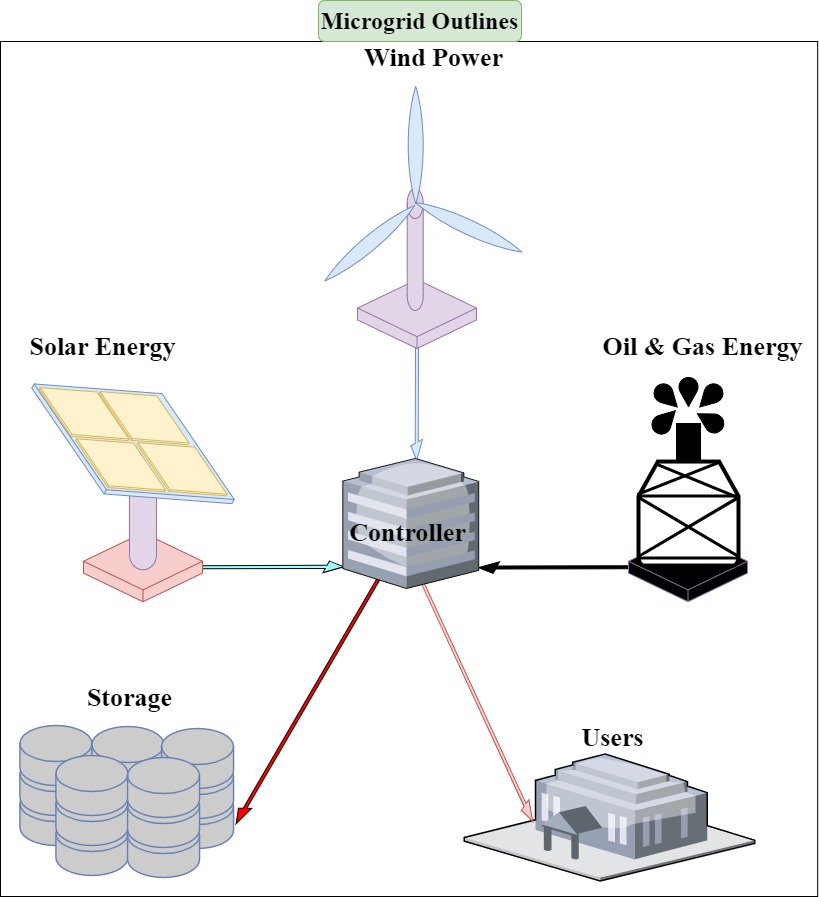}
\caption{{The microgrid components and its structure.}}
\label{fig20}
\end{figure}

In the microgrid ecosystem, load forecasting and the identification of anomalous generation patterns are integral to fulfilling demand requirements. The energy provision within such networks may derive from various sources, including diesel and solar energy, showcasing the system's versatility in harnessing power. \cite{abdelgawad2019comprehensive}. However, this form of distributed energy source is susceptible to fluctuations thereby presenting challenges in maintaining stable output power.Thus, conducting a thorough analysis of the patterns exhibited by the generated electricity is of paramount importance. On that note, the proliferation of diverse energy generation capabilities within microgrid configurations has markedly piqued the interest of a broader spectrum of stakeholders \cite{raza2015review}. In this research, the study delves into predictive analysis of volatile load demand and the detection of anomalies in power generation within microgrid infrastructures, aiming to enhance the operational efficiency and reliability of these systems. The proposed solution focuses on striking a balance between the demand and energy produced by the grid; this approach could further facilitate an effective management of grid energy even during the elevated usage in a typical scenarios such as warm seasons. Another benefit of the devised strategy is accurate load forecasting within the microgrids; this plays a crucial role in preventing unexpected outages, enhancing the reliability of power supply. Furthermore, the adoption of renewable energy sources in electricity generation significantly contributes to the reduction of CO2 emissions, promoting a more sustainable and environmentally friendly energy landscape. A typical representation of the micro grid is shown in Figure \ref{fig20}.
 
 As illustrated in the Figure \ref{fig20}, the microgrid integrates both renewable and non-renewable energy sources. Central to this system's efficacy is a sophisticated controller tasked with the precise apportionment of generated power. This includes a strategic division between direct residential supply and allocation to energy storage solutions, underpinning a nuanced approach to energy distribution and conservation. 
 
Artificial Intelligence (AI) has exhibited substantial potential in the analysis of time series data. This research is aimed at leveraging the structural advantages of well-established memory models, specifically the Gated Recurrent Unit (GRU) and attention mechanisms, to refine microgrid behavior predictions \cite{cahuantzi2023comparison}. Additionally, the study has explored the incorporation of convolution layers for extracting time-dependent features to further enhance the precision of microgrid predictions\cite{pan2023hybrid}. The proposed structure integrates a one-dimensional convolution function (Conv1d) and a Gated Recurrent Unit (GRU) to capture temporal features from the input time series. An attention layer further refines this process by highlighting the significance of specific time-dependent information extracted from the series, thereby enhancing the model's focus on relevant data for improved predictive accuracy. Finally, a Multilayer Perceptron (MLP) \cite{almeida2020multilayer} model is used for end to end regression and classification.To expedite the performance of the final model, the minimum number of units has been selected for both the GRU and MLP components.

The literature review in recent studies illustrates a broad exploration of AI techniques, particularly in Deep Learning and Machine Learning, for diagnosing anomalies and forecasting behaviors in microgrid systems. Various approaches, ranging from Feed-Forward Neural Networks (FNN), Recurrent Neural Networks (RNNs) including Long Short-Term Memory (LSTM) and Gated Recurrent Units (GRU), to more complex architectures like sequence-to-sequence models and Temporal Convolutional Networks (TCN), have been scrutinized \cite{10590520, ibrahim2022machine}. The study is notable for its extensive comparison of these models applied to real-world data, identifying specific models that excel in short-term and long-term load forecasting. Similarly, the research demonstrates the efficacy of integrating full wavelet packet transforms with Multi-Layer Perceptrons (MLP) for detailed electricity consumption forecasting during varying conditions, significantly outperforming standard neural network approaches\cite{el2020ensemble}.

Further, the application of machine learning and deep learning for electrical load prediction is elaborated in studies, which reveal the superior accuracy of LSTM-based models when optimized with relevant data, including weather variables\cite{chen2023short}. The innovative use of federated learning for energy forecasting showcases the potential for privacy-preserving models in residential energy consumption analysis\cite{hosseini2023privacy}. Moreover, the advancement in fault detection methods within microgrids, as discussed in recent studies, highlights the significant role of various machine learning classifiers, like Decision Trees and AdaBoost, in enhancing microgrid reliability. The collective findings from these studies underscore the evolving landscape of AI applications in microgrid monitoring and management, steering towards more efficient and reliable energy systems. This backdrop forms the foundation for our research, where we propose a new model combining convolution, GRU, and attention layers aimed at improving load forecasting precision and speed, addressing the challenges associated with training deep learning models for real-time microgrid applications.
 
The main focus of the proposed method is to extract effective information from the original time series using a shallow deep learning architecture. This research has three main contributions:
\begin{itemize}
\item Proposing a shallow network for load and abnormal behavior forecasting.
\item Comparing the performance of the proposed shallow deep learning model and traditional Machine Learning (ML) models.
\item Improving the performance of the proposed deep learning method using the fine-tuning of the learning rate and batch size. 
\end{itemize} 

This research paper continues as follows: Section II delves into the latest literature on load forecasting and the utilization trend in DL models, while section  II offers a statistical review of the datasets used. Section III, on the other hand, focuses on the evaluation of model parameters and algorithm structures for load forecasting and abnormal behavior prediction. The experimental results, along with the performance ratings on these models, are presented in Section IV. Finally, Section V gives a recap of the study's findings, limitations, and potential future directions.

\section{Methodology}

\subsection{Dataset}
The Microgrid Tariff Assessment Tool offers information on electricity consumption in rural sub-Saharan Africa, allowing users to examine the impact of system reliability on sizing and costs. It includes data on the levelized cost of electricity for diesel-only, solar-plus-storage, and solar-storage-diesel systems. The tool enables scenario comparison for locations in East, West, and Southern Africa, factoring in local geographical and economic parameters \cite{rangel2023optimisation}.
the tool allows comparison scenarios for three different locations in sub-Saharan Africa: East Africa (Lodwar, Kenya), West Africa (Accra, Ghana), and Southern Africa (Lusaka, Zambia). Each location has specific geographical and economic parameters tailored to the region, such as solar resource availability, diesel fuel prices, and distribution system costs. The statistical characteristics of the dataset are shown in Table \ref{tab:tableI}.
 
\begin{table}[h!]
    \centering
    \caption{Statistical characteristics of the dataset.}
    \label{tab:tableI}
    \setlength{\tabcolsep}{10pt} 
    \renewcommand{\arraystretch}{1.5} 
    \begin{tabular}{|l |r |r |}
        \hline
        \textbf{Data Feature} & \textbf{Average} & \textbf{Variance} \\  
        \hline
        PV (KW) & 70.84 & 8.45 \\          
        \hline
        Battery (KW) & 7.55 & 1.99 \\          
        \hline
        Battery (KWh) & 382.21 & 14.92 \\          
        \hline
        Generator (KW) & 18.55 & 5.46 \\          
        \hline
        PV capital cost & 237,795,130.20 & 15,034.17 \\          
        \hline
        PV O and M & 6,894,229.04 & 2,559.88 \\          
        \hline
        Battery capital cost & 284,208,250.30 & 12,202.27 \\          
        \hline
        Battery O and M & 17,504,910.43 & 3,115.69 \\          
        \hline
        Diesel capital cost & 6,325,958.59 & 3,040.68 \\          
        \hline
        Diesel O and M (KW) & 853,071.00 & 1,173.45 \\          
        \hline
        Diesel O and M (KWh) & 11,055,836.65 & 1,930.88 \\          
        \hline
        Fuel Cost & 1,038,589,968.04 & 35,854.60 \\          
        \hline
        REopt LLC & 442,926,489.30 & 74,736.26 \\          
        \hline
    \end{tabular}
\end{table}

In this research, two columns are considered as the main targets of the prediction. The first considered target in this research is the generator-produced power (kW) and abnormal behavior of the generator-produced power (kW). 

\subsection{Proposed Model}
\subsubsection{Convolution Layer}
This investigation utilizes a one-dimensional convolution layer to delineate the temporal associations among elements within a time series dataset\cite{kattenborn2021review}. The essence of this layer lies in its employment of mathematical convolutions, which process information across designated windows. The formula for computing the output of this convolution layer is shown in Equation \ref{eq1}.

\begin{equation}
\label{eq1}
y_{k}^{l} \ =\ b_{k}^{l} \ +\ {\displaystyle \sum _{i\ =\ 1}^{\ N_{l} -1} Conv_{1d}\left( w_{ik}^{l-1} *x_{_{i}}^{l-1}\right) \ \ \ \ \ \ \ \ }
\end{equation}
where the $_{_{i}}^{l-1}$ is defined as the input of the layer l-1, $b_{k}^{l}$
is defined as the bias of the $k_{th}$ neuron at layer l, $w_{ik}^{l-1}$ is the kernel from the 
$i_{th}$ neuron at layer $l-1$ to the $k_{th}$ neuron at layer l, $Conv_{1d}$ is the 1-dimensional convolution layer with zero padding function, $y_{k}^{l}$ is $k_{th}$ output of layer $l$. Utilizing a one-dimensional convolution layer presents a dual benefit of enabling the extraction of temporal features while concurrently facilitating the optimization of the end regression or classifier model\cite{kiranyaz20211d}. 
\subsubsection{Gated Recurrent Unit (GRU)}
\begin{figure}[htb]
\centering
\includegraphics[width=\columnwidth]{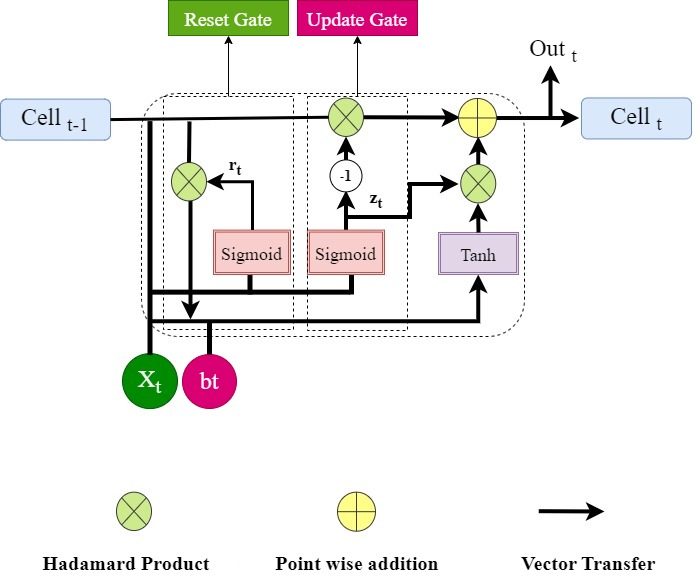}
\caption{Architecture of the GRU.}
\label{fig1}
\end{figure}

The Gated Recurrent Unit (GRU) is an improved version of the LSTM network. Different from LSTM, the GRU prioritizes passing only the most critical information while omitting non-vital data. It accomplishes this by employing fewer gates in its structure \cite{qian2023fault}. The architecture of the GRU layer is shown in Figure \ref{fig1}.
the equations for components of the GRU are shown as follows:
\begin{equation}
\label{eq2}
r_{t} \ =\ Sigmoid( X_{t} *W_{r} +W_{r} *H_{t-1} +b_{r})
\end{equation}
\begin{equation}
\label{eq3}
z_{t} \ =\ Sigmoid( X_{t} *W_{z} +W_{z} *H_{t-1} +b_{z})
\end{equation}
\begin{equation}
\label{eq4}
Out_{t} \ =\ ( 1-z_{t}) \ *h_{t-1} \ +z_{t} \ *( Tanh( X_{t} *W +W*r_{t} *h_{t-1} +b))
\end{equation}
The process of calculating the reset gate is illustrated in Equation \ref{eq2}, while the GRU unit's output computations are demonstrated in Equation \ref{eq3}. GRUs address the vanishing gradient descent issue in simple RNNs, improving training speed. They use update and reset gates to control information forwarded to the output, similar to LSTM cells but with one less gate, making them more efficient.
\subsubsection{Attention Layer}
\begin{figure}[htb]
\centering
\includegraphics[width=\columnwidth]{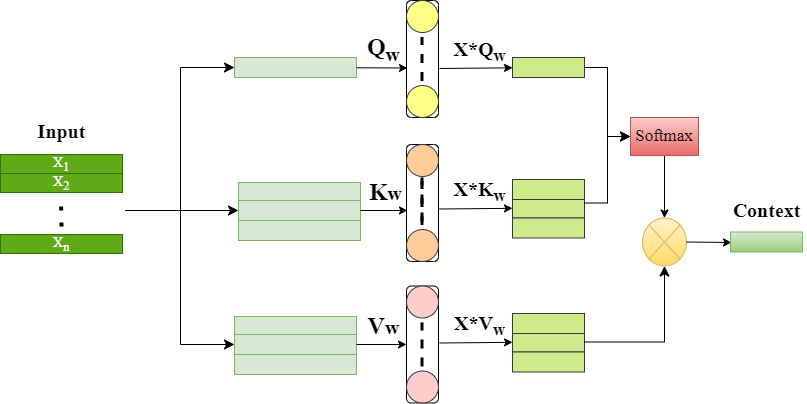}
\caption{Attention unit mechanism.}
\label{fig2}
\end{figure}

Central to each attention mechanism is the process to compute attention scores for every segment of the input sequence. These scores dictate the level of focus each part should receive during output generation. Typically, a softmax function is applied to calculate these attention scores, ensuring their sum equals one, thus facilitating a weighted importance across the sequence\cite{li2023deep}. The architecture of the attention layer is shown in Figure \ref{fig2}.
The attention weights are determined by assessing the likeness between the query and every key, using a measure of similarity such as dot product or cosine similarity. These weights are subsequently normalized using a softmax function to achieve a probability distribution. The importance of each occurrence of the query is then dictated by these weights \cite{guo2022attention}. The formula for calculating the emphasizing factor $\alpha$ is mentioned in Equation \ref{eq5}.
\begin{equation}
\label{eq5}
\alpha _{t,i} \ =\ \frac{e^{\ score( X*K_{w} ,X*Q_{w} ,)}}{{\displaystyle \sum _{i\ =\ 1}^{\ n}} e^{\ score( X*K_{w} ,X*Q_{w} ,)}}
\end{equation}
\subsubsection{Multilayer Perceptron (MLP)}
This study uses a Multilayer Perceptron (MLP) \cite{yan2019stat}, a commonly used architecture in artificial neural networks, as a regression model to forecast the final layer \cite{damacharlacovid}. Techniques such as the dropout layer and batch normalization are employed to prevent overfitting. The hidden layer within MLP modules consists of a combination of dropout and dense layers. The output layer's units depend on the grid's bus numbers \cite{sabzehgar2020solar}.


\subsection{Proposed structure}
\begin{figure*}[!t]
    \centering
    \includegraphics[width=\textwidth]{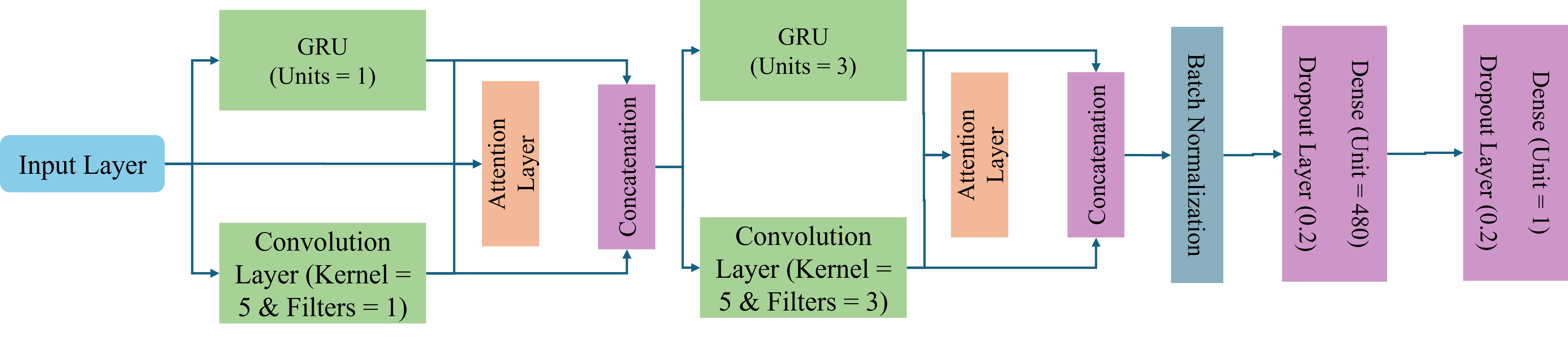}
    \caption{The proposed method architecture}
    \label{fig:fig3}
\end{figure*}

In this paper, the study introduces a novel framework combining convolutional, GRU, attention, and MLP layers for forecasting load or detecting anomalies within datasets. Initially, convolution and GRU layers process the input signal for time-sensitive patterns, while the attention layer identifies key insights from the GRU output. The outputs from these layers are merged and normalized, with the architecture being iteratively applied. The culmination of this process is the application of a multi-layer perceptron for final value prediction, with the architecture detailed in the accompanying figure. \ref{fig:fig3}.

\subsection{Baseline Models}

Similar to the reviewed research \cite{bouktif2018optimal}, this study conducts a comparative analysis of the proposed model against traditional ML models using an identical dataset. The evaluation encompasses KNN, SVM, tree-based methods, and Bayesian strategies to assess the proposed model's performance. Each model applies a unique analytical perspective to the classification and regression tasks, with subsequent sections delving into the structural composition of each model for a comprehensive understanding.
\subsubsection{KNN}
The K-Nearest-Neighbours (KNN) is a classification technique that uses the nearest neighbors of a data record to determine its category. Selecting the right value for 'k' in this method can be tricky \cite{cunningham2021k}. The KNN Model, a more efficient approach, automatically optimizes 'k', improving classification speed and accuracy while reducing data dependency, making it a viable alternative to traditional KNN classification \cite{SahuA2024}.

\subsubsection{SVM}
Support vector machines (SVMs) are popular supervised learning techniques used for data classification and information retrieval in large, multidimensional tasks \cite{awad2015support}. They are noted for their high accuracy in text categorization and pattern identification, especially with small and unbalanced training sets. While primarily used in binary classification, SVMs can also handle multi-class problems. Their theoretical and empirical benefits make them suitable for active learning

\subsubsection{Tree base algorithms}
A decision tree is a hierarchical tool used for data classification, boasting accurate, straightforward analysis \cite{zhao2020xgb}. XGBoost and AdaBoost are ensemble learning algorithms that combine weak learners into one effective learner. XGBoost is a quick, efficient, scalable method that reduces overfitting, whilst AdaBoost iteratively builds a strong classifier, focusing on hard-to-categorize cases. Both are widely used in machine learning applications due to their improved predicted accuracy \cite{navada2011overview}.

\subsubsection{Beysian Methods}
The Naive Bayes algorithm, based on Bayes' Theorem, is effective for classifying data such as tumors using gene expression profiling. Despite assuming each property is independent, it can predict classes based on assigned posterior class probabilities  \cite{zampieri2021using}. Highly parallelizable and flexible, it's suitable for large data analytics and applicable to fields like statistics and bioinformatics. A Bayesian probabilistic model must be built to use this method.

\section{Experimental Results And Analysis}

This study conducted experiments using 80\% of the dataset for training and 20\% for testing, with training capped at 10,000 iterations and discontinued if no improvement was seen after 300 epochs. The initial learning rate was set at 0.001 and was reduced threefold in the absence of performance gains. Evaluations were performed using a V-100 GPU and an Intel Cori@7 processor, employing MAE and RMSE as the key metrics.The formulae for calculating the MAE and RMSE is mentioned as follows:
\begin{gather}
Mean\ Absolut\ Error\ ( MAE) \ \ =\\
\ \frac{\sum _{i=1}^{N} |Predicted_{i} \ -\ Real_{i} \ |}{N} \notag
\end{gather}
\begin{gather}
Root\ Mean\ Squared\ Error\ ( RMSE) \ \ \ = \notag\\
\sqrt{\frac{\sum _{i=1}^{N}\left( Predicted_{i} \ -\ Real_{i} \ \right)^{2}}{N}}
\end{gather}
\begin{equation}
Accuracy = \frac{TP+TN}{TP+FP+TN+FN}
\end{equation}
\\
\begin{equation}
Precision = \frac{TP}{TP+FP}
\end{equation}
\\
$TP$, $TN$, $FP$, and $FN$ are abbreviations for true positive, true negative, false positive, and false negative predictions, respectively.

\begin{table}[h!]
    \centering
    \caption{Comparison between various DL models for load forecasting.}
    \label{tab:tableII}
    \setlength{\tabcolsep}{5pt} 
    \renewcommand{\arraystretch}{1.3} 
    \begin{tabular}{|l |r |r |r|}
        \hline
        \textbf{Model} & \textbf{MAE} & \textbf{RMSE} & \textbf{R2-score} \\  
        \hline
        CNN-GRU-Attention & 0.39 & 0.28 & 98.89 \\          
        \hline
        SVR & 1.18 & 3.81 & 80.54 \\          
        \hline
        KNN & 0.41 & 0.82 & 95.70 \\          
        \hline
        XGB & 0.59 & 1.03 & 94.74 \\          
        \hline
        Bayesian Ridge & 0.49 & 0.44 & 97.74 \\          
        \hline
        RF & 0.25 & 0.48 & 97.53 \\          
        \hline
        CNN-GRU-Attention (Kenya) & 0.38 & 0.27 & 98.88 \\          
        \hline
        SVR (Kenya) & 1.31 & 4.63 & 76.42 \\          
        \hline
        KNN (Kenya) & 0.61 & 0.39 & 96.84 \\          
        \hline
        XGB (Kenya) & 0.64 & 1.03 & 94.57 \\          
        \hline
        Bayesian Ridge (Kenya) & 0.45 & 0.51 & 97.66 \\          
        \hline
        RF (Kenya) & 0.45 & 0.51 & 97.66 \\          
        \hline
    \end{tabular}
\end{table}

\begin{figure}[htb]
\centering
\includegraphics[width=1\columnwidth]{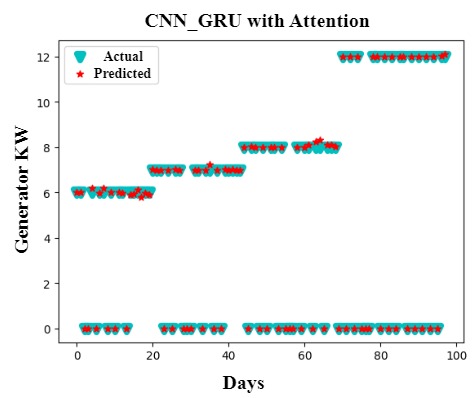}
\caption{The performance of the model for predicted versus real value.}
\label{fig6}
\end{figure}

The evaluated results for forecasting the load by the proposed DL model amd ML models has shown in Table \ref{tab:tableII}. As shown in Table \ref{tab:tableII}, the performance of the proposed method is superior to ML models. The proposed model performance vs real value for the predicted instances is shown in Figure \ref{fig6}.

The analysis across various machine learning models reveals competent accuracy in identifying instances with zero output. Nevertheless, the models' efficacy in predicting values generated by the generator falls short. Specifically, the Support Vector Regression (SVR) model from the experiment results shows suboptimal performance for predicting peak generation values around 12kW. Conversely, the Random Forest (RF) model excels at predicting zero-output instances but struggles with accurately forecasting generated energy levels of 8kW, highlighting a nuanced challenge in energy production prediction.
\\
The proposed methodologies exhibit robust performance in forecasting non-generative states and various generation levels. A distinctive strength of the model lies in its precision for predicting instances with zero generated power. The model's versatility is further demonstrated by assessing its performance under zero-generation conditions and other scenarios through a modification in its final layer for classification purposes. The efficacy of this model, particularly in predicting zero-output states, is illustrated in the accompanying figure. 

\begin{figure}[htb]
\centering
\includegraphics[width=\columnwidth]{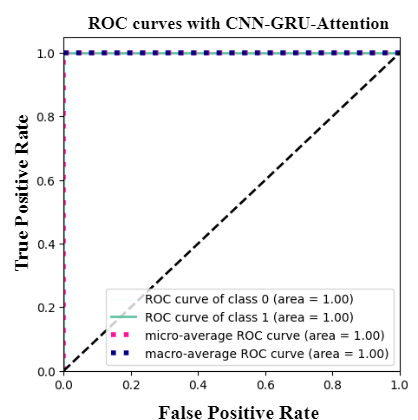}
\caption{ROC Curve Showing Performance of Final CNN-GRU Model}
\label{fig7}
\end{figure}

\begin{figure}[htb]
\centering
\includegraphics[width=1\columnwidth]{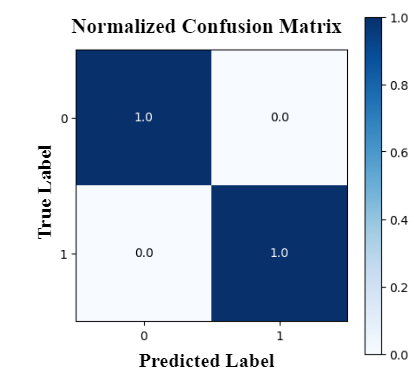}
\caption{True Table Showing Performance of Final CNN-GRU Model}
\label{fig8}
\end{figure}

Figures ~\ref{fig7} and \ref{fig8} illustrate that the model's capability in classifying non-generative states shows promising results, highlighting its potential in analyzing fluctuations in power production. This suggests the model could be valuable in predicting power outputs, even when faced with significant oscillations.
\section{Discussion}
\begin{figure}[htb]
\centering
\includegraphics[width=\columnwidth]{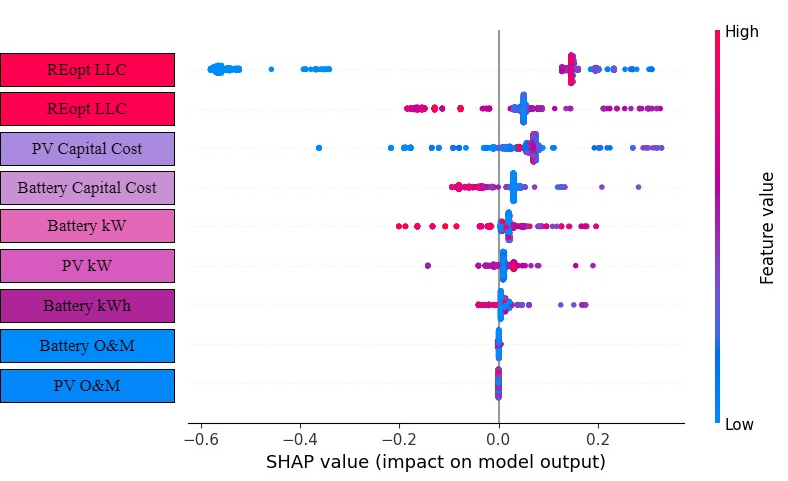}
\caption{The importance of each feature based on shapely value.}
\label{fig9}
\end{figure}

This study employs a synergistic approach utilizing CNN, GRU, and attention mechanisms to forecast microgrid operations and identify anomalies, extracting temporal patterns from various energy sources for predictive load management. It employs an attention layer to highlight crucial features, validated against the Micro-grid Tariff Assessment Tool dataset, achieving approximately 99.9\% accuracy in zero state predictions and load forecasting accuracy with metrics of 0.39 MAE, 0.28 MSE, and an r2-score of 98.89\%. The significance of features on predictions is assessed using Shapley values.
The results of shapely values for each feature is shown in Figure \ref{fig9}.

Shapley values are a concept from cooperative game theory and are used as a method of attributing a fair reward to players based on their contribution to the game. Within the context of predictive modeling, these values offer a quantitative metric to assess the relative importance of each feature, enabling a nuanced understanding of their impact on model outcomes.
As shown in Figure \ref{fig9}, "Battery O\&M" and "Reopt LLC" are the two most important features for forecasting the generated power. Also, the "PV O\&M" and "Battery O\&M" are the least important features for abnormal behavior detection and generated load forecasting.  
\\
The model exhibits near-optimal performance in precise load forecasting and anomaly detection, attributed to the optimized number of neurons and units within the MLP and GRU layers, respectively. This optimization renders the method relatively shallow, ensuring rapid response times that rival those of machine learning models and surpass the base algorithm in test phases. Consequently, this approach is well-suited for real-time applications in load forecasting and the detection of abnormal behaviors. 
\\
Future work will focus on enhancing the model's accuracy in predicting microgrid behaviors by exploring various optimization techniques to fine-tune the hyperparameters,thereby striving to achieve superior performance of the proposed approach. 

\section{Conclusion}
A microgrid system consists of many diverse renewable and non-renewable energy sources that feed into either a standalone system or a utility grid. The energy produced varies significantly due to the range of sources, like solar and diesel, leading to fluctuations between no power and full power generation. These fluctuations can result in abnormal behavior in the microgrid. Incorporating a mix of renewable and non-renewable sources, this study devises a predictive model for microgrid dynamics, utilizing convolution and GRU layers for time-sensitive data extraction, emphasized by an attention layer. An MLP model refines the prediction and abnormality detection. Evaluated with the Micro-grid Tariff Assessment Tool dataset, this method demonstrated accuracy with a 0.39 MAE, 0.28 RMSE, and 98.89\% r2-score in load forecasting, and nearly 99.9\% accuracy in zero power state prediction, showcasing its effectiveness in microgrid behavior forecasting.
 
\printbibliography
\end{document}